\documentclass[runningheads,a4paper]{llncs}
\usepackage{amsmath,amssymb}
\usepackage{graphicx}
\graphicspath{ {images/} }
\usepackage{multirow}
\usepackage{url}
\usepackage{verbatim}
\usepackage[colorlinks=true,citecolor=blue]{hyperref}

\urldef{\mailsa}\path|rcha9612@uni.sydney.edu.au, etot5316@uni.sydney.edu.au, schawla@qf.org.qa|

\newcommand{\keywords}[1]{\par\addvspace\baselineskip
\noindent\keywordname\enspace\ignorespaces#1}
\usepackage{float}
\usepackage{subfigure}
\usepackage{multirow}
\usepackage{soul,color,colortbl}
\usepackage{xcolor}
\usepackage{booktabs}
\usepackage{upgreek}
\usepackage[noend]{algpseudocode}
\usepackage[ruled,vlined,linesnumbered]{algorithm2e}

\usepackage{booktabs} 
\usepackage[english]{babel}
\usepackage[utf8]{inputenc}
\usepackage{amssymb}

\newcommand{\A}{\mathbf{A}}
\newcommand{\B}{\mathbf{B}}
\newcommand{\E}{\mathbf{E}}
\newcommand{\I}{\mathbf{I}}
\newcommand{\J}{\mathbf{J}}
\newcommand{\N}{\mathbf{N}}
\newcommand{\R}{\mathbf{R}}
\renewcommand{\S}{\mathbf{S}}
\newcommand{\T}{\mathbf{T}}
\newcommand{\U}{\mathbf{U}}
\newcommand{\V}{\mathbf{V}}
\newcommand{\W}{\mathbf{W}}
\newcommand{\X}{\mathbf{X}}
\newcommand{\Z}{\mathbf{Z}}

\newcommand{\Real}{\mathbb{R}}
\newcommand{\sanjay}[1]{\hl{\footnote{\hl{Sanjay: #1}}}}

\let\oldFootnote\footnote
\newcommand\nextToken\relax

\renewcommand\footnote[1]{%
    \oldFootnote{#1}\futurelet\nextToken\isFootnote}

\newcommand\isFootnote{%
    \ifx\footnote\nextToken\textsuperscript{,}\fi}

\mainmatter  

\title{Group Anomaly Detection using Deep Generative Models}

\titlerunning{Group Anomaly Detection using Deep Generative Models}

\usepackage[symbol]{footmisc}
 \author{Raghavendra Chalapathy\inst{1} \footnote {Equal Contribution \label{foot1}} \and Edward Toth\inst{2 }   \footref{foot1} \and Sanjay Chawla\inst{3}}

 \authorrunning{Chalapathy, Toth and Chawla}

\institute{The University of Sydney and Capital Markets CRC
\and
School of Information Technologies, The University of Sydney
\and
Qatar Computing Research Institute, HBKU
}

\toctitle{Lecture Notes in Computer Science}
\tocauthor{Authors' Instructions}
\begin{document}
\maketitle
\begin{abstract}

Unlike conventional anomaly detection research that focuses on point anomalies, our goal is to detect anomalous collections of individual data points. In particular, we perform group anomaly detection (GAD) with an emphasis on irregular group distributions (e.g. irregular mixtures of image pixels). GAD is an important task in detecting unusual and anomalous phenomena in  real-world applications such as high energy particle physics, social media and medical imaging. In this paper, we take a generative approach by proposing deep generative models: Adversarial autoencoder (AAE) and variational autoencoder (VAE) for group anomaly detection. Both AAE and VAE detect group anomalies using point-wise input data where group memberships are known a priori. We conduct extensive experiments to evaluate our models on real world datasets. The empirical results demonstrate that our approach is effective and robust in detecting group anomalies.


\end{abstract}

\section{Anomaly detection: motivation and challenges}

 Group anomaly detection (GAD) is an important part of data analysis for many interesting group applications.
Pointwise anomaly detection  focuses on the study of individual data instances that do not conform with the expected pattern in a dataset. With the increasing availability of multifaceted information, GAD research has recently explored datasets involving groups or collections of observations.  Many pointwise anomaly detection methods cannot detect a variety of different deviations that are evident in group datasets.   For example,  Muandet et al. \cite{OCSMM} possibly discover Higgs bosons as a group of collision events in high energy particle physics whereas pointwise methods are unable to distinguish this anomalous behavior. Detecting group anomalies require more specialized techniques for robustly differentiating group behaviors.




GAD aims to identify groups that deviate from the regular group pattern.  Generally, a group consists of a collection of two or more points and group behaviors are more adequately described by a greater number of observations. A point-based anomalous group is a collection of  individual pointwise anomalies that deviate from the expected pattern. It is more difficult to detect distribution-based group anomalies where points are seemingly regular however their collective behavior is anomalous.  It is also possible to characterize group anomalies by certain properties and subsequently apply pointwise anomaly detection methods. In image applications, a distribution-based anomalous group has an irregular mixture of visual features compared to the expected group pattern.   

\begin{figure}[h]
\includegraphics[scale=0.38,
trim=0.5cm 1cm 0.5cm 0.8cm]
{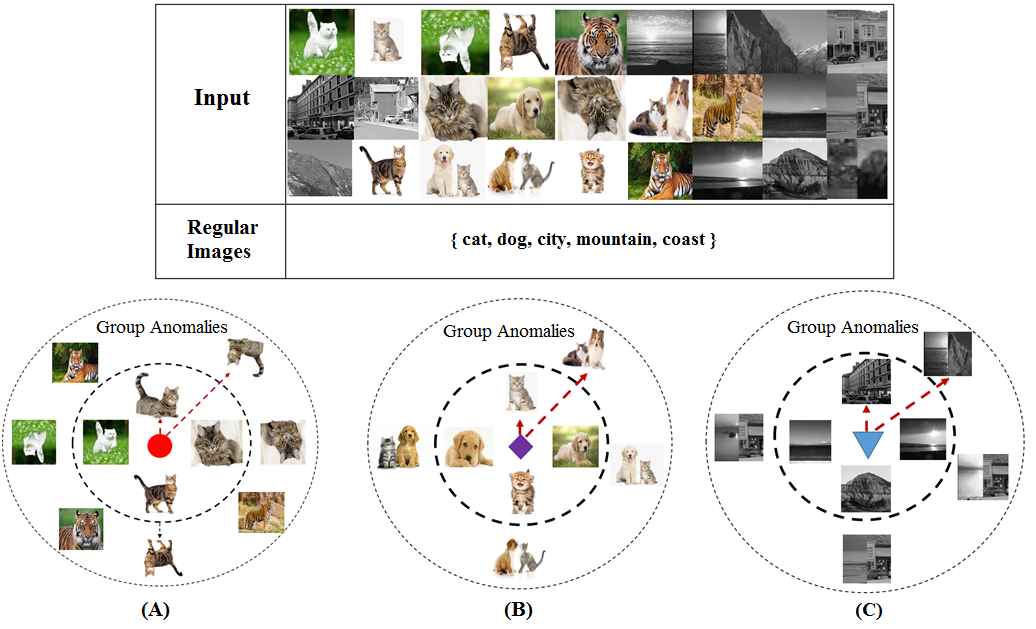}
\centering
\caption{ Examples of point-based and distribution-based group anomalies in various image applications. The expected group behavior represents images in the inner concentric circle while the outer circle contains images that are group anomalies. 
}
\label{fig:GAD}
\end{figure}

In this paper, an image is modeled as a group of pixels or visual features (e.g. whiskers in a cat image).  
Figure \ref{fig:GAD} illustrates various examples of point-based and distribution-based group anomalies where the inner circle contains images exhibiting regular behaviors whereas images in the outer circle represent group anomalies. For example, the innermost circle in plot (A) contains regular images of cats whereas the outer circle portrays point-based group anomalies where  individual features of tigers (such as stripes) are point anomalies.   On the other hand, plot (A) also illustrates distribution-based group anomalies where rotated cat images  have anomalous distributions of   visual cat features.  In plot (B),   distribution-based group anomalies are irregular mixtures of cats and dogs in a single image while plot (C) depicts anomalous  images stitched  from different scene categories of cities, mountains or coastlines. 
Our image data experiments will mainly focus on detecting group anomalies in these scenarios.

Even though the GAD problem may seem like a straightforward comparison of group observations, many complications and challenges arise. As there is a dependency between the location of pixels in a high-dimensional space, appropriate features in an image may be difficult to extract. For effective detection of anomalous images, an adequate description of images is required  for model training.  Complications in images potentially arise such as low resolution, poor illumination intensity, different viewing angles, scaling and rotations of images.  Like other anomaly detection applications, ground truth labels are also usually unavailable for training or evaluation purposes.
A number of pre-processing and extraction techniques can be applied as solutions to different aspects of these challenges.

In order to detect  distribution-based group anomalies in various image applications, we propose using  deep generative models (DGMs). 
The main contributions of this paper are: 
\begin{itemize}
\item We formulate DGMs for the problem of detecting group anomalies using a group reference function. 
\item Although deep generative models have been applied in various image applications, they have not been applied to the GAD problem. 
\item  A variety of experiments are performed on both  synthetic and real-world datasets to demonstrate the effectiveness of  deep generative models for detecting group anomalies as compared to other GAD techniques. 
\vspace{1mm}
\end{itemize}

The rest of the paper is organized as follows. An overview of related work is provided (Section~\ref{sec:related}) and preliminaries for understanding approaches for detecting  group anomalies are also described (Section~\ref{sec:preliminaries}). 
We formulate our problem and then proceed to elaborate on our proposed solution that involves deep generative models (Section~\ref{sec:method}). 
Our experimental setup and key results are presented in Section~ \ref{sec:experiment-setup}  and Section ~\ref{sec:experiment-results} respectively.
Finally, Section~\ref{sec:conclusion} provides a summary of our findings as well as recommends future directions for GAD research. 

\section{Background and related work on group anomaly detection}
\label{sec:related}

GAD applications are emerging areas of research where most state-of-the-art techniques have been more recently developed.  While group anomalies are briefly discussed in anomaly detection surveys such as Chandola et al. \cite{Chandola} and Austin \cite{Hodge},  Xiong \cite{Collective} provides a more detailed description of current state-of-the-art GAD methods.  Yu et al. \cite{SurveySocialMedia} further reviews GAD techniques where group structures are not previously known however  clusters are inferred based on additional information of pairwise relationships between data instances. 
We explore  group anomalies when group memberships are known a priori such as in image applications. 

Previous studies on image anomaly detection 
can be understood in terms of group anomalies. 
Quellec et al.  \cite{mammo} examine mammographic images  where point-based group anomalies represent potentially cancerous regions. Perera and Patel \cite{chairs} learn features from a collection of images containing regular chair objects and detect point-based group anomalies where chairs have abnormal shapes, colors and other irregular characteristics. On the other hand, regular categories in Xiong et al. \cite{FGM}  represent scene images such as inside city, mountain or coast  and distribution-based group anomalies are stitched images with a mixture of different scene categories. At a pixel level,   Xiong et al. \cite{MGM} apply GAD methods to detect anomalous galaxy clusters with irregular proportions of RGB pixels. We emphasize detecting distribution-based group anomalies rather than point-based anomalies in our subsequent  image applications.

The discovery of group anomalies is of interest to a number of diverse domains.  
  Muandet et al.	\cite{OCSMM} investigate GAD for physical phenomena in high energy particle physics where Higgs bosons are observed as slight excesses in a collection of collision events rather than individual  events. Xiong et al. \cite{FGM} analyze a fluid dynamics application where a group anomaly represents unusual vorticity and turbulence in  fluid motion.   In topic modeling,  Soleimani and Miller \cite{ATD} characterize documents by topics and anomalous clusters of documents are discovered by their irregular  topic mixtures.  By incorporating additional information from pairwise connection data, Yu et al. \cite{GLAD} find potentially irregular communities of co-authors in various research communities.
Thus there are many GAD application other than image anomaly detection.


A  related discipline to image anomaly detection is video anomaly detection where many deep learning architectures have been applied. 
 Sultani  et al.  \cite{survideos1} detect real-world anomalies such as burglary, fighting, vandalism and so on from  CCTV footage using deep learning methods.  
 In a review, Kiran et al. \cite{survideos2} compare DGMs with different  convolution architectures for  video anomaly detection applications. 
Recent work ~\cite{schlegl2017unsupervised,xu2018unsupervised,an2015variational} illustrate the effectiveness of generative models for high-dimensional anomaly detection. Although, there are existing works that have applied deep generative models in image related applications, they have not been formulated as a GAD problem. We leverage  autoencoders for DGMs 
to detect group anomalies in a variety of data experiments. 



\section{Preliminaries}
\label{sec:preliminaries}
In this section, a summary of state-of-the-art techniques for detecting group anomalies is provided.
We also assess strengths and weaknesses of existing  models, compared with the proposed deep generative models.

\subsection{Mixture of Gaussian Mixture Models (MGMM) }
\label{sec:mgmm}

A hierarchical generative approach MGMM is proposed by Xiong et al. \cite{MGM} for detecting group anomalies. The data generating process in MGM  assumes that each group follow  a Gaussian mixture where more than one regular mixture proportion is possible.  For example, an image is a distribution over visual features such as paws and whiskers from a cat image and each image is categorized into possible regular behaviors or genres 
(e.g. dogs or cats).  An anomalous group is then characterized by an irregular mixture of visual features such as a cat and dog in a single image.
MGM is useful for distinguishing multiple types of group behaviors however poor results are obtained when group observations do  not appropriately follow the assumed generative process.

\subsection{One-Class Support Measure Machines (OCSMM)}
\label{sec:ocsmm}

 Muandet et al. \cite{OCSMM} propose OCSMM to maximize the margin that separates regular class of group behaviors from anomalous groups.    Each group is firstly characterized by a mean embedding function then group representations   are separated by a parameterized hyperplane.
 OCSMM is able to classify groups as regular or anomalous behaviors however careful parameter selection is required in order to effectively detect group anomalies.


\subsection{One-Class Support Vector Machines (OCSVM) }
\label{sec:ocsvm}

 If group distributions are reduced and characterized by a single value then OCSVM from Sch{\"o}lkopf et al. \cite{OCSVM}  can be applied to the GAD problem. OCSVM separates data points using a parametrized hyperplane similar to OCSMM. 
 OCSVM requires additional pre-processing to convert groups of visual features into pointwise observations. We follow a bag of features approach in Azhar et al. \cite{SIFT-OCSVM}, where $k$-means is applied to visual image features and centroids are clustered into histogram intervals before implementing OCSVM.  OCSVM is a popular pointwise anomaly detection method however it may not accurately capture group anomalies if the initial group characterizations are inadequate.




\subsection{Deep generative models for anomaly detection}
\label{sec:adversarialAE}
This section describes the mathematical background of deep generative models that will be applied for detecting group anomalies. 
\subsubsection{Autoencoders:}
\label{sec:autoencoders}
An autoencoder is trained to learn reconstructions that are close to its original input. The autoencoder consists of  encoder $f_\phi$ to embed the input to latent or hidden representation and  decoder $g_\psi$ which reconstructs the input from hidden representation. The reconstruction loss of an autoencoder is defined as the squared error between the input $G_{m}$  and output $\hat G_{m}$ given by

\begin{equation}
{ L_r(G_{m},\hat G_{m} )} = ||{ G_m - \hat G_m }||^2  \hspace{0.5cm}  \mbox{where } G_m \in \mathbb{R}^{N \times V}
\label{eqn:aeloss}
\end{equation}
Autoencoders leverage reconstruction error as an anomaly score where data points with significantly high errors are considered to be anomalies.

\subsubsection{Variational Autoencoders (VAE):}
\label{sec:autoencoders}
Variational autoencoder (VAE)~\cite{Kingma2013} are generative analogues to the standard deterministic autoencoder.
VAE impose constraint while inferring  latent variable $z$. The hidden latent codes produced by encoder $f_\phi$ is constrained to follow prior data  distribution  $P(G_m)$. The core idea of VAE is to infer $P(z)$
from $P(z|G_m)$ using Variational Inference (VI) technique given by

\begin{equation}
{ L(G_m,\hat G_m)} = { L_r(G_m,\hat G_m)} + KL(f_\phi(z|x)\, || \, g_\psi(z))  
\label{eqn:vaeloss}
\end{equation}

In order to optimize the Kullback–Leibler (KL) divergence, a simple reparameterization trick is applied; instead of the encoder embedding a real-valued vector, it creates a vector of means $\boldsymbol \mu$ and a vector of standard deviations $\boldsymbol  \sigma$. Now a new sample that replicates the data distribution $P(G_m)$ can be generated from learned parameters ($\boldsymbol \mu$,$\boldsymbol \sigma$)  and input this latent representation $z$ through the decoder $g_\psi$ to reconstruct the original group observations. VAE utilizes reconstruction probabilities~\cite{an2015variational} or reconstruction error to compute anomaly scores.

\subsubsection{Adversarial Autoencoders (AAE):}
\label{sec:aae}
One of the main limitations of VAE is lack of closed form analytical solution for integral of the KL divergence term except for few distributions. Adversarial autoencoders (AAE)~\cite{makhzani2015adversarial} avoid using the KL divergence by adopting adversarial learning, to learn broader set of distributions as priors for the latent code.
The training procedure for this architecture  is performed using an adversarial autoencoder consisting of  encoder $f_\phi$  and decoder $g_\psi$. Firstly a latent representation $z$ is created according to generator network $f_\phi(z|G_m)$ ,and the decoder reconstructs the input $\hat G_m$ from $z$. The weights of encoder $f_\phi$ and decoder $g_\psi$ are updated by backpropogating the reconstruction loss between $\hat G_m$ and $G_m$.
Secondly the discriminator receives  $z$ distributed as  $f_\phi(z|G_m)$ and  $z'$ sampled from the true prior  $P(z)$ to compute the score assigned to each ($D(z)$ and $D(z')$). The loss incurred is minimized by backpropagating through the discriminator to update its weights. The loss function for autoencoder (or generator) $L_G$ is composed of the reconstruction error along with the loss for  discriminator $L_D$  where
\begin{equation}
\begin{aligned}
{L_G} = \frac{1}{M'} \sum_{m=1}^{M'} \log D(z_m) \mbox{ \;  and \; } L_D = -\frac{1}{M'} \sum_{m=1}^{M'} \big [\log D(z'_m)+ \log(1- D(z_m)) \big ]
\end{aligned}
\label{eqn:aaeloss}
\end{equation}
where $M'$ is the minibatch size while $z$ represents the latent code generated by encoder and $z'$ is a sample from the true prior $P(z)$.


\section{Problem and Model Formulation }
\label{sec:method}
\subsubsection{Problem Definition:}
Suppose we observe a set of groups $\mathcal{G} = \big\{  {\bf G}_m \big\} _{ m=1 }^M  $ where the $m$th group contains $N_m$ observations with 

\begin{equation}
{\bf G}_m = \big( X_{ij}\big) \in \mathbb{R}^{N_m \times V} \label{Eqn:Group}
\end{equation}
where the total number of individual observations is $N=\sum_{m=1}^M N_m$.

In GAD, the behavior or properties of the $m$th group is captured by a characterization function denoted by $f:  \mathbb{R}^{N_m \times V} \to \mathbb{R}^{D}$ where $D$ is the dimensionality on the transformed feature space. After a characterization function is applied to a training dataset,  group information is combined using an aggregation function $g: \mathbb{R}^{M \times D} \to \mathbb{R}^{D}$.  A group reference is composition of characterization and aggregation functions on the input groups with 
\begin{align}
\mathcal{G}^{(ref)} = g \Big[ \big\{ f({\bf G}_{m} ) \big\}_{m=1}^M \Big]
\label{eqn:Gref}
\end{align}
Then a distance metric $d(\cdot,\cdot) \ge 0  $ is applied to measure the deviation of a particular group from the group reference function. The distance score $  d\Big(\mathcal{G}^{(ref)}  , {\bf G}_{m} \Big )$  quantifies the deviance of the $m$th group from the expected group pattern where larger values are associated with more anomalous groups. 
Group anomalies are effectively detected when characterization function $f$ and aggregation function $g$  respectively capture properties of group distributions and appropriately combine information into a group reference. For example in an variational autoencoder setting, an encoder function $f$ characterizes mean and standard deviation  of group distributions with $f\big( {\bf G}_m \big) = ( {\mu}_m,{\sigma}_m)   $ for $ m = 1,2,\dots,M $. 
Obtaining the  group reference using an aggregation function in deep generative models is more challenging where VAE and AAE are further explained in Algorithm \ref{algo:gadVae}. 




\vspace{4mm}
\begin{algorithm}[H]
\DontPrintSemicolon
\SetAlgoLined
\SetKwInOut{Input}{Input}\SetKwInOut{Output}{Output}
\Input{ Groups $\mathcal{G} = \big\{  {\bf G}_m \big \} _{ m=1 }^M  $  where  ${\bf G}_m = \big( X_{ij}\big) \in \mathbb{R}^{N_m \times V} $
}
\BlankLine
\Output{Group anomaly scores \textbf{S} for input groups $\mathcal{G}$ }
\BlankLine
Train AAE and VAE to obtain encoder $f_\phi$ and decoder $g_\psi$ on inputs  $\mathcal{G}$    \;
\BlankLine
  \Begin{
        \Switch{C}{
            \Case{(VAE)}{
              \For{(m = 1 to M)}{
    			\BlankLine
      				$(\mu_m,\sigma_m) = f_\phi(z|G_m)$\;  
                }
           $(\mu,\sigma) = \frac{1}{M}\sum_{m=1}^{M}      (\mu_m,\sigma_m$)\;
               \BlankLine
    draw a sample from $z \sim \mathcal{N}(\mu,\,\sigma)$\;
            }
            \Case{(AAE)}{
          draw a random latent representation $z \sim f_\phi(z|\mathcal{G})$                         }
              reconstruct sample using decoder $\mathcal{G}^{(ref)}=    g_\psi(\mathcal{G}|z)$\;   
          \For{(m = 1 to M)}{
               compute the score
        ${ s}_m =d\Big(\mathcal{G}^{(ref)}, {\bf G}_{m}   \Big ) $

        \BlankLine
    }
     sort scores in descending order  
     \textbf{S}=$ \{s_{(M)} >\dots>s_{(1)} \}$\;
    groups that are furthest from $\mathcal{G}^{(ref)}$ are more anomalous.\;   
        }
        \textbf{return S}
    }

\caption{Group anomaly detection using deep generative models}
\label{algo:gadVae}
\end{algorithm}

\subsection{Training the model}
\label{sec:training}
The variational and adversarial autoencoder are trained according to the objective function given in Equation ~(\ref{eqn:vaeloss}), (\ref{eqn:aaeloss}) respectively. The objective functions of DGMs are optimized using standard backpropogation algorithm. Given known  group memberships, AAE is fully trained on input groups to obtain a  representative group reference $\mathcal{G}^{(ref)}$ from Equation \ref{eqn:Gref}. While in case of VAE, $\mathcal{G}^{(ref)}$ is obtained by drawing samples using mean and standard deviation parameters that are inferred using VAE as illustrated in Algorithm~\ref{algo:gadVae}.

\subsection{Predicting with the model}
In order to identify  group anomalies, the distance of a group from  the group reference   $\mathcal{G}^{(ref)}$ is computed. The output scores are sorted according to descending order where groups that are furthest from $\mathcal{G}^{(ref)}$ are considered anomalous. One convenient property of DGMs is that the anomaly detector will be inductive, i.e.  it can generalize to unseen data points. One can interpret the model as learning a robust representation of  group distribution. An appropriate characterization allows for the  accurate detection  where any unseen  observations  either  lie within the reference group manifold or they are deviate from the expected group pattern. 

\section{Experimental setup}
\label{sec:experiment-setup}
In this section we show the empirical effectiveness of deep generative models over the state-of-the-art methods on real-world data. Our primary focus will be on non-trivial image datasets, although our method is applicable in any context where autoencoders are useful e.g.\ speech, text.

\subsection{Methods compared}
We compare our proposed  technique using deep generative models (DGMs)
with the following state-of-the art methods for group anomaly detection:
\let\labelitemi\labelitemii
\begin{itemize}

	\item \textbf{Mixture of Gaussian Mixture Model (MGMM)}, as per~\cite{MGM}.
	\item \textbf{One-Class Support Measure Machines (OCSMM)}, as per~\cite{OCSMM}.
	\item \textbf{One-Class Support Vector Machines (OCSVM)}, as per~\cite{OCSVM}.
 	\item \textbf{Variational Autoencoder (VAE)}~\cite{doersch2016tutorial}, as per Equation (\ref{eqn:vaeloss}).
	\item \textbf{Adversarial Autoencoder (AAE)}~\cite{makhzani2015adversarial}, as per Equation (\ref{eqn:aaeloss}).
\end{itemize}

We used Keras~\cite{chollet2015keras}, TensorFlow \cite{abadi2016tensorflow} for the implementation of AAE and  VAE~\footnote{\url{https://github.com/raghavchalapathy/gad}}. MGM \footnote{\url{https://www.cs.cmu.edu/~lxiong/gad/gad.html}}, OCSMM
\footnote{\url{https://github.com/jorjasso/SMDD-group-anomaly-detection}}
and OCSVM 
\footnote{\url{
https://github.com/cjlin1/libsvm}}
are implemented
publicly available code

\subsection{Datasets}
We compare all methods on the following datasets: 
\begin{itemize}

	\item {\tt cifar-10}~\cite{krizhevsky2009learning} consisting of $32\times32$ color images over 10 classes with 6000 images per class.
    \item {\tt scene} images from different categories following Xiong et al. ~\cite{xiong2011group}.

\end{itemize}
The real-world data experiments using {\tt cifar-10}  and {\tt scene}  data are visually summarized in Figure~\ref{fig:GAD} . 




\subsection{Parameter Selection} 
We now briefly discuss the model and parameter selection for applying techniques in GAD applications. 
A pre-processing stage is required for state-of-the-art GAD methods when dealing with images 
where feature extraction methods such as SIFT \cite{sift} or HOG \cite{hog}  represent images as a collection of visual features. 
In MGM, the number of regular group behaviors $T$ and number of Gaussian mixtures $L$ are selected using information criteria.  The kernel bandwidth smoothing parameter  in OCSMM \cite{OCSMM} is chosen as $  \mbox{median}\big\{ || {\bf G}_{m,i} -{\bf G}_{l,j} ||^2 \big\} $ for all $i,j \in \{1,2,\dots,N_m \}$ and $m,l \in {1,2,\dots,M}$ where $ {\bf G}_{m,i} $ represents the $i$th random vector in the $m$th group.  In addition, the parameter for expected  proportions of anomalies in  OCSMM and OCSVM is set to the true value in the respective datasets however this is not required for other techniques.

When applying VAE and AAE, there are four existing  network parameters that require careful selection;
(a) number of convolutional filters, (b) filter size, (c) strides of convolution operation and (d) activation function. We tuned via grid search additional hyper-parameters, including the number of hidden-layer nodes $H \in \{3, 64, 128\}$, and regularization  $\lambda$ within range ${[0, 100]}$. The learning drop-out rates and regularization parameter $\mu$ were sampled from a uniform distribution in the range $[0.05, 0.1]$. The embedding and initial weight matrices are all sampled from uniform distribution within range $[-1, 1]$.

\section{Experimental results}
\label{sec:experiment-results}

In this section, we explore a variety of experiments for group anomaly detection. As anomaly detection is an unsupervised learning problem, model evaluation is highly challenging. Since we known the ground truth labels of the group anomalies that are injected into the real-world image datasets. The performance of  deep generative models is evaluated against state-of-the-art GAD methods using area under the precision-recall curve (AUPRC) and area under the ROC curve (AUROC) metrics. AUPRC and AUROC quantify the  ranking performance of different methods where the former metric is more appropriate under class imbalanced datasets ~\cite{Davis:2006}.

\subsection{Synthetic Data: Rotated Gaussian distribution }
We generate synthetic dataset where
regular behavior consists of  bivariate Gaussian samples while anomalous groups have are rotated covariance structure. 

More specifically,  $M=500$ regular group distributions have correlation $\rho =0.7$ while 50 anomalous groups have correlation $\rho =-0.7$.  The mean vectors are randomly sampled from uniform distributions 
,while  covariances of group distributions are
 \begin{equation}
  \boldsymbol\Sigma_m=\left\{
  \begin{array}{@{}ll@{}}

  \;
\begin{pmatrix}
     0.2 & 0.14 \\
  0.14 & 0.2
  \end{pmatrix}, & m=1,2,\dots,500 \\[5mm]
  \;  \small \begin{pmatrix}
     0.2 & -0.14 \\
  -0.14 & 0.2
  \end{pmatrix}, &m=501,502,\dots,550 \end{array}\right.
    \label{changecovariance}
\end{equation}
with each group having $N_m = 1536$ observations.


\textbf{Parameter settings}:
GAD methods are applied on the raw data with various parameter settings.
MGM is trained with $T=1$ regular scene types and $L=3$ as the number of Gaussian mixtures. The expected proportion of group anomalies as true proportion in OCSMM and OCSVM is set to  $\nu$ = 50/M where $M= 550$ or  $M= 5050$. In addition, OCSVM is applied by treating each Gaussian distribution as a single high-dimensional observation.

\textbf{Results}: Table \ref{tbl:syn} illustrates the results of detecting distribution-based group anomalies for different group sizes. For smaller group sizes $M= 550$, state-of-the-art GAD methods achieve a higher performance  than deep generative models however for a larger training set with $M= 5050$, deep generative models achieve the highest performance. This conveys that deep generative models require larger number of group observations in order to train an appropriate model. 

\begin{table*}
    \centering
    \scalebox{1}{
    \begin{tabular}{|c|c|c|c|c|c|c|}
        \hline
        \multirow{2}{*}{Methods} &
        \multicolumn{2}{c|}{\bf { M=550}} &
            \multicolumn{2}{c|}{\bf { M=5050}}
        \\
        \cline{2-5}
        &AUPRC & AUROC &AUPRC & AUROC
          \\
        \hline
        AAE &$0.9060$ &$0.5000$ &\cellcolor{gray!25}$1.0000$ &\cellcolor{gray!25}$1.0000$ \\
        VAE &$0.9001$ &$0.5003$ &\cellcolor{gray!25}$1.0000$&\cellcolor{gray!25}$1.0000$ \\
        \midrule
        MGM&\cellcolor{gray!25}$0.9781$ &\cellcolor{gray!25}$0.8180$
           &$0.9978$ &$0.8221$\\
        OCSMM&$0.9426$ &$0.6097$
                  &$0.9943$ &$0.6295$
        \\
        OCSVM&$0.9211$ &$0.5008$&$0.9898$ &$0.5310$
        \\
        \hline
\end{tabular}}
       \vspace{2mm}
        \caption{Task results for detecting rotated Gaussian distributions in synthetic datasets, the first two rows contains deep generative models and the later techniques are state-of-the-art GAD methods. The highest performances are shaded in gray.  }
    \label{tbl:syn}
\end{table*}

\subsection{Detecting tigers within cat images }
\label{sec:tigerDitect}
Firstly we explore the detection of point-based group anomalies (or image anomalies) by combining 5000  images of cats and 50 images of tigers. The images of tigers were obtained from Pixabay~\cite{pixabayImages} and rescaled to match the image dimension of cats in {\tt cifar-10} dataset. From Figure~\ref{fig:GAD}, cat images are considered as regular behavior while  characteristics of tiger images are point anomalies. The goal is to correctly detect all images of tigers in an unsupervised manner.

\textbf{Parameter settings}:
In this experiment, HOG extracts visual features as inputs for GAD methods. MGM is trained with $T=1$ regular cat type and $L=3$ as the number of mixtures. Parameters in OCSMM and OCSVM are set to  $\nu$ = 50/5050 and OCSVM is applied with $k$-means ($k=40$). For the deep learning models, following the success of the Batch Normalization architecture~\cite{ioffe2015batch} and Exponential Linear Units~\cite{clevert2015fast}, we have found that convolutional+batch-normalization+elu layers provide a better representation of convolutional filters. Hence, in this experiment the autoencoder of both AAE and VAE adopts four layers of (conv-batch-normalization-elu) in the encoder part and four layers of  (conv-batch-normalization-elu) in the decoder portion of the network. AAE network parameters such as (number of filter, filter size, strides) are chosen to be (16,3,1) for first and second layers and (32,3,1) for third and fourth layers of both encoder and decoder layers. The middle hidden layer size is set to be same as rank $K = 64$ and the model is trained using Adam~\cite{kingma2014adam}.
The decoding layer uses sigmoid function in order to capture the nonlinearity characteristics from  latent representations produced by the hidden layer.


\subsection{Discovering rotated entities }
Secondly we explore the detection of distribution-based group anomalies in terms of a rotated images by examining 5000  images of cats and 50 images of rotated cats. As illustrated in Figure~\ref{fig:GAD}, images of rotated cats have anomalous distributions  compared to regular images of cats . In this scenario,  the goal is to detect all rotated cats in an unsupervised manner.

\textbf{Parameter settings}:
In this experiment, HOG extracts visual features for this rotated group anomaly experiment because SIFT features are rotation invariant. MGM is trained with $T=1$ regular cat type and $L=3$ as the number of mixtures while  $k$-means ($k=40$) is applied to the SIFT features and then OCSVM is computed with  parameters $\nu$ is set to the true  proportion  of anomalies. For the deep generative models, the network parameters follows similar settings as described in previous experiment of detecting tigers within cats images.

\subsection{Detecting cats and dogs }
We further investigate the detection of distribution-based group anomalies in terms of a cats with dogs. The constructed dataset consists of 5050 images; 2500 single cats, 2500 single dogs and 50 images of  cats and dogs together. The 50 images of cats and dogs together were obtained from Pixabay~\cite{pixabayImages} and rescaled to match the image dimension of cats and dogs present in {\tt cifar-10} dataset. 
Images with a single cat or dog are considered as regular groups while images with both cats and dogs are distributed-based group anomalies.  In this scenario,  the goal is to detect all images with irregular mixtures of cats and dogs in an unsupervised manner.

\textbf{Parameter settings}:
In this experiment, HOG extracts visual features as inputs for GAD methods. 
MGM is trained with $T=2$ regular cat type and $L=3$ as the number of mixtures while  OCSVM is applied with $k$-means ($k=30$).
For the deep generative models, the parameter settings follows setup as described in Section~\ref{sec:tigerDitect}. 
\begin{figure}[!t]
    \centering
     \subfigure[ Tigers within cat images from {\tt cifar-10} dataset.]{\includegraphics[scale=0.50]
{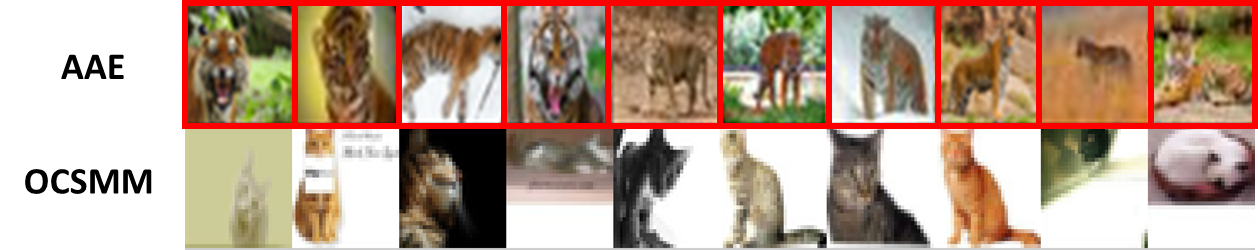}}
\subfigure[Images of cats and  dogs within single cat and dog images using {\tt cifar-10} dataset. ]{\includegraphics[scale=0.5]
{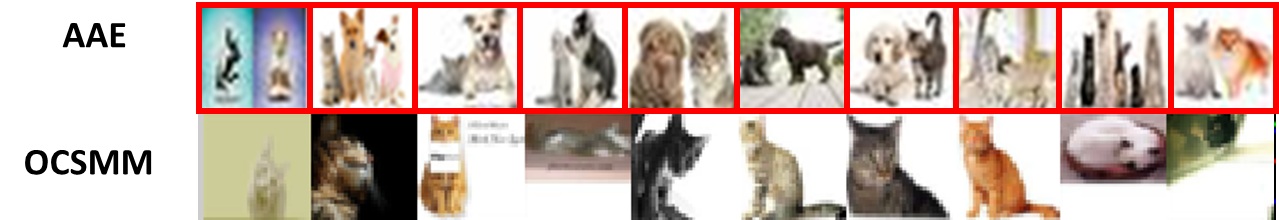}}
    \caption{
    Top 10 anomalous images are presented from the best deep learning and GAD method. Red boxes around images  represent correctly detected group anomalies as shown in (a) and (b). AAE correctly detects all group anomalies in the {\tt cifar-10} dataset containing tigers as well as the experiment involving images with cats and dogs. While OCSMM consistently identifies similar cat images as the most anomalous images.
    }
    \label{fig:results-cifar}
\end{figure}

\subsection{Detecting stitched scene images}
We present results on scene image dataset where 
 100 images originated from each category “inside city”, “mountain" and “coast”. An additional 66 stitched anomalies contain images with two scene categories. 
For example, distribution-based group anomalies may  represent images with half coast and half city street view. These anomalies are challenging since they have the same local features as regular images however as a collection, they are anomalous. 

\textbf{Parameter settings}:
State-of-the-art GAD methods utilize SIFT feature extraction for visual image features in this experiment . MGM is trained with $T=3$ regular scene types and $L=4$ Gaussian mixtures while 
OCSVM is applied with $k$-means ($k=10$). 
The scene images are rescaled from  dimension $256 \times 256$ to $32 \times 32$ to enable reuse the same architecture followed in previous experiments.
The parameter settings for both AAE and VAE  follows setup as described in Section~\ref{sec:tigerDitect}.

\begin{figure}[!t]
    \centering
    \subfigure[ Rotated cats  amongst regular cats in the {\tt cifar-10} dataset. ]{\includegraphics[scale=0.47]{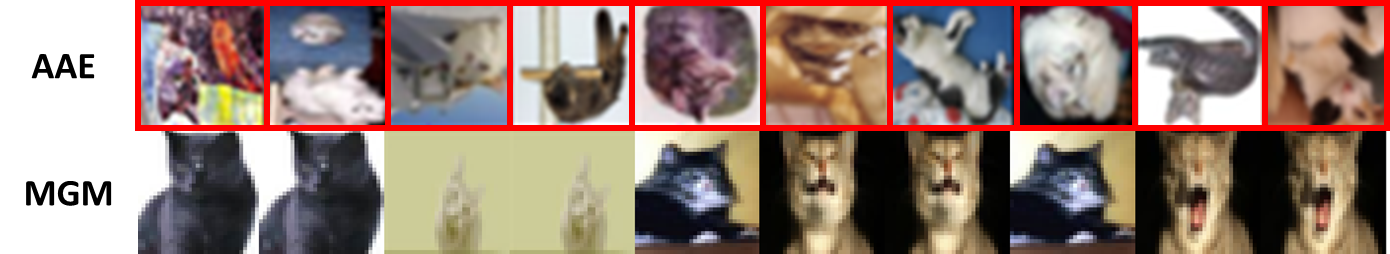}}
    \subfigure[Stitched Images amongst the {\tt scene} dataset.  ]{\includegraphics[scale=0.5]{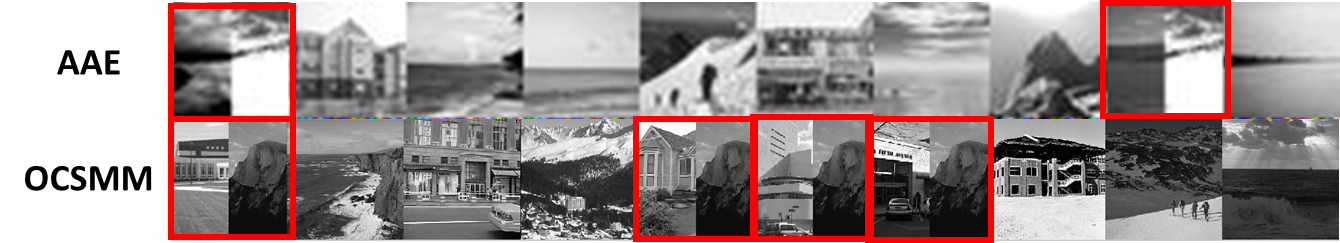}}
    \caption{Top 10 anomalous images are presented where  red boxes outlining images  represent true group anomalies in the given datasets. AAE performs well in (a) however does not effectively detect group anomalies in (b).  MGM is unable to detect rotated cats based on extracted  HOG features while OSCMM is able to detect group anomalies in the {\tt scene} dataset.}
    \label{fig:results-rotatedcatsandscene}
\end{figure}

\subsection{ Results Summary and Discussion}

Table  ~\ref{tablesummary} summarizes the detection performance of 
 deep generative models and existing  GAD methods on a variety of datasets.  AAE achieves the highest detection performance in  experiments (except for {\tt scene} data) as illustrated in Figure \ref{fig:results-cifar} and ~\ref{fig:results-rotatedcatsandscene}. 
 Similar to the results in the synthetic dataset, deep generative models have a significantly worse performance when the group size is small such as in the {\tt scene} dataset with $M=366$ compared to $M=5050$ groups in other experiments. DGMs are effective in detecting group anomalies for  larger training set. 


\begin{table*}
        \setlength{\tabcolsep}{1.2mm}
    \scalebox{1}{
    \begin{tabular}{|c|c|c|c|c|c|c|c|c|c|c|}

        \hline
        \multirow{2}{*}{Methods} &
        \multicolumn{2}{c|}{\bf {\small    Tigers }}  &
                \multicolumn{2}{c|}{\bf {\small Rotated Cats}} &
                           \multicolumn{2}{c|}{\bf {\small Cats and Dogs }} &
                                  \multicolumn{2}{c|}{\bf {\small Scene }} \\
        \cline{2-9}
     &AUPRC & AUROC        & AUPRC & AUROC      &AUPRC & AUROC        & AUPRC & AUROC
        \\
        \hline
        \multirow{4}{*}{}
        AAE& $0.9449$ &\cellcolor{gray!25} $0.9906$
         &\cellcolor{gray!25}$1.0000$ &\cellcolor{gray!25}$1.0000$
            &\cellcolor{gray!25}$1.0000$ &\cellcolor{gray!25}$1.0000$
        & \cellcolor{gray!25} 0.9449 & 0.5906
        \\
        VAE&$0.9786$ &$0.9092$ 
        &$0.9999$ &$0.9999$
               &$0.9998$ &$0.9999$
                  &$0.8786$ &$0.3092$
        \\
        \midrule
        MGM & $0.9881$ & $0.5740$
        &$0.9919$ &$0.6240$ 
        &0.9906 & 0.5377
               &0.8835 & 0.6639
        \\
        OCSMM
             &\cellcolor{gray!25} $0.9941$ &$0.6461$
        &$0.9917$ &$0.6128$
        & 0.9930 & 0.5876
             &$0.9140$ &\cellcolor{gray!25} $0.7162$
        \\
        OCSVM
        & 0.9909 & 0.5474
        &$0.9894$ &$0.5568$
        & 0.9916 & 0.5549
        & 0.8650 & 0.5733
        \\
        \hline
\end{tabular}}
       \vspace{2mm}
        \caption{Summary of results for various data experiments where first two rows contains deep generative models and the later techniques are state-of-the-art GAD methods. The best performances are shaded in gray.}
    \label{tablesummary}
\end{table*}

{\bf Comparison of training times:}
\label{sec:runtime}
We add a final remark about applying the proposed deep generative models on  GAD problems in terms of computational time and training efficiency.  
For example, including the time taken to calculate SIFT features on the small-scale {\tt scene} dataset, MGMM takes 42.8 seconds for training, 3.74 minutes to train OCSMM and 27.9 seconds for OCSVM. In comparison, the computational times for our AAE and VAE are  6.5 minutes and 8.5 minutes respectively. All the experiments involving deep learning models were conducted on a MacBook Pro equipped with an Intel Core i7 at 2.2 GHz, 16 GB of RAM (DDR3 1600 MHz). The ability to leverage recent advances in deep learning as part of our optimization (e.g. training models on a GPU) is  a salient feature of our approach.
We also note that while MGM and OCSMM method are faster to train on small-scale datasets, they suffer from at least $O(N^2)$ complexity for the total number of observations $N$. 
It is plausible that one could leverage recent advances in fast approximations of kernel methods~\cite{Lopez-Paz:2014} for OCSMM and studying these would be of interest in future work. 

\section{Conclusion}
\label{sec:conclusion}
Group anomaly detection is a challenging area of research especially when dealing with complex group distributions such as image data. In order to detect  group anomalies in various image applications, we clearly formulate deep generative models (DGMs)  for  detecting distribution-based group anomalies.   
 DGMs outperform state-of-the-art GAD techniques in many experiments involving both synthetic and real-world image datasets.  However, DGMs also require a large number of group observations for model training.
To the best of our knowledge, this is the first paper to formulate and apply deep generative models to the problem of detecting group anomalies. A future direction for research involves using recurrent neural networks  to detect  temporal changes in a group of time series.

\bibliographystyle{splncs03}
\bibliography{MyBibFile,outlier}

\end{document}